\documentclass[letterpaper]{article} 
\usepackage{aaai23}  
\usepackage{times}  
\usepackage{helvet}  
\usepackage{courier}  
\usepackage[hyphens]{url}  
\usepackage{graphicx} 
\usepackage{natbib}  
\usepackage{caption} 
\usepackage{algorithm}
\usepackage[noend]{algpseudocode}

\usepackage{newfloat}
\usepackage{listings}
\DeclareCaptionStyle{ruled}{labelfont=normalfont,labelsep=colon,strut=off} 
\floatstyle{ruled}
\newfloat{listing}{tb}{lst}{}
\floatname{listing}{Listing}

\usepackage{hyperref}
\usepackage{bbm}
\usepackage{amsmath}
\usepackage{amssymb}
\usepackage{xcolor}
\usepackage{vector}
\usepackage{cleveref}
\usepackage{siunitx}
\usepackage{booktabs}

\newcommand{\obsdist}{\ensuremath{\mathcal{Z}}}

\newcommand{\ra}[1]{\renewcommand{\arraystretch}{#1}}
\newcolumntype{R}{>{$}r<{$}}

\nocopyright
\title{Online Planning for Constrained POMDPs with \\ Continuous Spaces through Dual Ascent
}
\author {
    Arec Jamgochian, 
    Anthony Corso, 
    Mykel J. Kochenderfer 
}
\affiliations {
    Stanford University, Department of Aeronautics and Astronautics\\
    \{arec, acorso, mykel\}@stanford.edu
}

\begin{document}

\maketitle

\begin{abstract}
Rather than augmenting rewards with penalties for undesired behavior, Constrained Partially Observable Markov Decision Processes (CPOMDPs) plan safely by imposing inviolable hard constraint value budgets. Previous work performing online planning for CPOMDPs has only been applied to discrete action and observation spaces. In this work, we propose algorithms for online CPOMDP planning for continuous state, action, and observation spaces by combining dual ascent with progressive widening. 
We empirically compare the effectiveness of our proposed algorithms on continuous CPOMDPs that model both toy and real-world safety-critical problems.
Additionally, we compare against the use of online solvers for continuous unconstrained POMDPs that scalarize cost constraints into rewards, and investigate the effect of optimistic cost propagation.

\end{abstract}

\section{Introduction}

Partially observable Markov decision processes (POMDPs) provide a mathematical framework for planning under uncertainty~\cite{sondik1978optimal,kochenderfer2022algorithms}. An optimal policy for a POMDP maximizes the long-term reward that an agent accumulates while considering uncertainty from the agent's state and dynamics. Planning, however, is often multi-objective, as agents will typically trade-off between maximizing multiple rewards and minimizing multiple costs. Though this multi-objectivity can be handled explicitly~\cite{roijers2013survey}, often times, multiple objectives are scalarized into a single reward function and penalties are captured through soft constraints. The drawback to this approach, however, is the need to define the parameters that weight the rewards and costs.

Constrained POMDPs (CPOMDPs) model penalties through hard constraint budgets that must be satisfied, but are often harder to solve than POMDPs with scalarized reward functions. Policies for CPOMDPs with discrete state, action, and observation spaces can be generated offline through point-based value iteration~\cite{kim2011point}, with locally-approximate linear programming~\cite{poupart2015approximate}, or projected gradient ascent on finite-state controllers~\cite{wray2022scalable}. Additional work develops an online receding horizon controller for CPOMDPs with large state spaces by combining Monte Carlo Tree Search (MCTS)~\cite{silver2010monte} with dual ascent to guarantee constraint satisfaction~\cite{lee2018monte}. However, this method is limited to discrete state and action spaces.

In this work, we extend MCTS with dual ascent to algorithms that leverage double progressive widening~\cite{sunberg2018online,couetoux2011continuous} in order to develop online solvers for CPOMDPs with large or continuous state, action, and observation spaces. Specifically we extend three continuous POMDP solvers (POMCP-DPW, POMCPOW and PFT-DPW)~\cite{sunberg2018online} to create the constrained versions (CPOMCP-DPW, CPOMCPOW, and CPFT-DPW). In our experiments, we a) compare our method against an unconstrained solver using reward scalarization, b) compare the algorithmic design choice of using simulated versus minimal cost propagation, and c) compare the rewards and costs accumulated by our three constrained solvers.  
\section{Background}
\subsubsection{POMDPs}
Formally, a POMDP is defined by the 7-tuple $(\mathcal{S},\mathcal{A},\mathcal{O},\mathcal{T},\mathcal{Z},\mathcal{R},\gamma)$ consisting respectively of state, action, and observation spaces, a transition model mapping states and actions to a distribution over resultant states, an observation model mapping an underlying transition to a distribution over emitted observations, a reward function mapping an underlying state transition to an instantaneous reward, and a discount factor. An agent policy maps an initial state distribution and a history of actions and observations to an instantaneous action. An optimal policy acts to maximize expected discounted reward~\cite{sondik1978optimal,kochenderfer2022algorithms}. 

Offline POMDP planning algorithms \cite{spaan2005perseus,kurniawati2008sarsop} yield compact policies that act from any history but are typically limited to relatively small state, action, and observation spaces. Online algorithms yield good actions from immediate histories during execution~\cite{ross2008online}. \citet{silver2010monte} apply Monte-Carlo Tree Search (MCTS) over histories to plan online in POMDPs with large state spaces (POMCP). 
Progressive widening is a technique for slowly expanding the number of children in a search tree when the multiplicity of possible child nodes is high or infinite (i.e. in continuous spaces)~\cite{couetoux2011continuous,chaslot2008progressive}. 
\citet{sunberg2018online} apply double progressive widening (DPW) to extend POMCP planning to large and continuous action and observation spaces. Additional work considers methods for selecting new actions when progressively widening using a space-filling metric \cite{lim2021voronoi} or Expected Improvement exploration \cite{mern2021bayesian}. \citet{wu2021adaptive} improve performance by merging similar observation branches. 

\subsubsection{Constrained planning}
Constrained POMDPs augment the POMDP tuple with a cost function $\mathcal{C}$ that maps each state transition to a vector of instantaneous costs, and a cost budget vector $\hat{\vect{c}}$ that the expected discounted cost returns must satisfy. An optimal CPOMDP policy $\pi$ maximizes expected discounted reward subject to hard cost constraints:
\begin{align}
\max_\pi & \ V_R^\pi(b_0)= \mathbb{E}_\pi \left[\sum_{t=0}^\infty \gamma^t R(b_t,a_t) \mid b_0 \right] \\
\text{s.t.   } & V_{C_k}^\pi(b_0)= \mathbb{E}_\pi \left[\sum_{t=0}^\infty \gamma^t C_k(b_t,a_t) \mid b_0 \right] \leq \hat{c}_k \ \forall \ k\text{,}
\end{align}
where $b_0$ is the initial state distribution, and belief-based reward and cost functions return the expected reward and costs from transitions from states in those beliefs. 

\citet{altman1999constrained} overview fully-observable Constrained Markov Decision Processes (CMDPs), while \citet{piunovskiy2000constrained} solve them offline using dynamic programming on a state space augmented with a constraint admissibility heuristic. Early offline CPOMDP solvers use an alpha vector formulation for value and perform cost and reward backups~\cite{isom2008piecewise} or augment the state space with cost-to-go and perform point-based value iteration~\cite{kim2011point}. CALP~\cite{poupart2015approximate} uses a fixed set of reachable beliefs to perform locally-approximate value iteration quickly by solving a linear program, and iterates in order to expand the set of reachable beliefs necessary to represent a good policy.~\citet{walraven2018column} improve upon this by leveraging column generation. 
\citet{wray2022scalable} represent a CPOMDP policy with a finite state controller and learns its parameters offline through projected gradient ascent.

To generate good actions online,~\citet{undurti2010online} perform look-ahead search up to a fixed depth while using a conservative constraint-minimizing policy learned offline to estimate the cost at leaf nodes and prune unsafe branches. 
More recently, CC-POMCP performs CPOMDP planning by combining POMCP with dual ascent ~\cite{lee2018monte}. Besides tracking cost values, CC-POMCP maintains estimates for Lagrange multipliers for the CPOMDP problem that it uses to guide search. Between search queries, CC-POMCP updates Lagrange multipliers using constraint violations. CC-POMCP outperforms CALP in large state spaces. 

\section{Approach}

The Lagrangian of the constrained POMDP planning problem can be formulated as 
\begin{equation}
\max_\pi\min_{\vect{\lambda}\geq0} [V_R^\pi(b_0)-\vect{\lambda}^\top(V_{\vect{C}}^\pi(b_0)-\hat{\vect{c}})].
\end{equation}
CC-POMCP~\cite{lee2018monte} optimizes this objective directly by interweaving optimization for $\pi$ using POMCP and optimization of $\vect{\lambda}$ using dual ascent. The planning procedure for CC-POMCP is summarized in the first procedure in Listing 1, with lines 6 and 7 depicting the dual ascent phase. During policy optimization, actions are always chosen with respect to the value of the Lagrangian (lines 10 and 11). 
The \texttt{StochasticPolicy} procedure (not shown) builds a stochastic policy of actions that are $\nu$-close to maximal Lagrangian action-value estimate by solving a linear program.

\begin{algorithm}[t]
    \floatname{algorithm}{Listing}
    \caption{Common procedures} \label{alg:common}
    \begin{algorithmic}[1]
        \Procedure{Plan}{$b$}
            \State $\vect{\lambda} \gets \vect{\lambda}_0$
            \For{$i \in 1:n$}
                \State $s \gets \text{sample from }b$
                \State $\Call{Simulate}{s, b, d_\text{max}}$
                \State $a \sim \Call{GreedyPolicy}{b,0,0}$
                \State $\vect{\lambda} \gets [\vect{\lambda} + \alpha_i (\vect{Q}_C(ba)-\hat{\vect{c}})]^+$
            \EndFor
            \State $\textbf{return } \Call{GreedyPolicy}{b,0,\nu}$
        \EndProcedure
        \Procedure {GreedyPolicy}{$h,\kappa,\nu$}
            \State $Q_\lambda(ba) := Q(ba)-\vect{\lambda}^\top \vect{Q}_C(ba) + \kappa \sqrt{\frac{\log N(h)}{N(ha)}}$
            \State $\textbf{return } \Call{StochasticPolicy}{Q_\lambda(ba),\nu}$
        \EndProcedure
        
        \Procedure {ActionProgWiden}{$h$}
            \If{$|C(h)| \leq k_a N(h)^{\alpha_a}$}
                \State $a \gets \Call{NextAction}{h}$
                \State $C(h) \gets C(h) \cup \{a\}$
            \EndIf
            \State $\pi \gets \Call{GreedyPolicy}{h,c,\nu}$
            \State $\textbf{return } 
            \text{sample from } \pi$
        \EndProcedure

    \end{algorithmic}
\end{algorithm}

One significant shortcoming of using POMCP for policy optimization is that it only admits discrete action and observation spaces. In this work, we address this shortcoming using progressive widening, in which a new child node is only added to parent node $p$ when the number of children nodes $|C(p)|\leq kN(p)^\alpha$, where $N(p)$ is the number of total visits to $p$, and $k$ and $\alpha$ are tree-shaping hyperparameters.

Progressively widening can be straightforwardly applied to CC-POMCP. In \Cref{sec:cpomcpdpw}, we provide an outline of the resulting algorithm, CPOMCP-DPW. However, as noted by~\citet{sunberg2018online}, progressively widening the observation searches will lead to particle collapse as each observation node after the first step would only hold a single state. 
To alleviate this, they propose POMCPOW to iteratively build up particle beliefs at each observation node and PFT-DPW perform belief tree search in continuous spaces using particle filter beliefs. 
In \Cref{alg:pomcpow,alg:cpft}, these approaches for POMDP planning in continuous spaces are combined with dual ascent in order to perform CPOMDP planning in continuous spaces. We note that these algorithms are amenable to methods that implement better choices for actions~\cite{mern2021bayesian,lim2021voronoi} or combine similar observation nodes~\cite{wu2021adaptive}

\setcounter{algorithm}{0}
\begin{algorithm}[t]
    \caption{CPOMCPOW} \label{alg:pomcpow}
    \begin{algorithmic}[1]
        \Procedure {Simulate}{$s$, $h$, $d$}        
            \If{$d = 0$}
                \State \textbf{return} $0$
            \EndIf
            \State $a \gets \Call{ActionProgWiden}{h}$
            \State $s',o,r,\vect{c} \gets G(s,a)$
            \If{$|C(ha)| \leq k_o N(ha)^{\alpha_o}$}
                \State $M(hao) \gets M(hao) + 1$
            \Else
                \State $o \gets \text{select } o \in C(ha) \text{ w.p. } \frac{M(hao)}{\sum_{o} M(hao)}$
            \EndIf
            \State $\text{append } s' \text{ to } B(hao)$ \label{lin:insert}
            \State $\text{append } \obsdist(o \mid s, a, s') \text{ to } W(hao)$ \label{lin:weight}
            \If{$o \notin C(ha)$} \Comment{new node}
                \State $C(ha) \gets C(ha) \cup \{o\}$
                \State $V', \vect{C}' \gets \Call{Rollout}{s', hao, d-1}$
            \Else
                \State $s' \gets \text{select } B(hao)[i] \text{ w.p. } \frac{W(hao)[i]}{\sum_{j=1}^m W(hao)[j]}$ \label{lin:sample}
                \State $r \gets R(s,a,s')$
                \State $\vect{c} \gets C(s,a,s')$
                \State $V', \vect{C}' \gets \Call{Simulate}{s', hao, d-1}$
            \EndIf
            \State $V \gets r + \gamma V'$
            \State $\vect{C} \gets \vect{c} + \gamma \vect{C}'$
            \State $N(h) \gets N(h)+1$
            \State $N(ha) \gets N(ha)+1$
            \State $Q(ha) \gets Q(ha) + \frac{V - Q(ha)}{N(ha)}$
            \State $\vect{Q}_C(ha) \gets \vect{Q}_C(ha) + \frac{\vect{C} - \vect{Q}_C(ha)}{N(ha)}$
            \State $\bar{\vect{c}}(ha) \gets \bar{\vect{c}}(ha) + \frac{\vect{c} - \bar{\vect{c}}(ha)}{N(ha)}$
            \If{returnMinimalCost}
                \State $\vect{C} \gets \underset{a \in C(h)}{\min}\, \vect{Q}_C(ha)$
            \EndIf
            \State \textbf{return} $V, \vect{C}$
        \EndProcedure
    \end{algorithmic}
\end{algorithm}
\begin{algorithm}
    \caption{CPFT-DPW} \label{alg:cpft}
    \begin{algorithmic}[1]
        \Procedure {Simulate}{$\cdot$, $b$, $d$}        
            \If{$d = 0$}
                \State \textbf{return} $0$
            \EndIf
            \State $a \gets \Call{ActionProgWiden}{b}$
            \If{$|C(ba)| \leq k_o N(ba)^{\alpha_o}$}
                \State $b',r,\vect{c} \gets G_\text{PF($m$)}(b,a)$
                \State $C(ba) \gets C(ba) \cup \{(b',r,\vect{c})\}$
                \State $V',\vect{C}' \gets \Call{Rollout}{b', d-1}$
            \Else
                \State $b', r, \vect{c} \gets \text{sample uniformly from } C(ba)$
                \State $V',\vect{C}' \gets \Call{Simulate}{\cdot, b', d-1}$
            \EndIf
            \State $V \gets r + \gamma V'$
            \State $\vect{C} \gets \vect{c} + \gamma \vect{C}'$
            \State $N(b) \gets N(b)+1$
            \State $N(ba) \gets N(ba)+1$
            \State $Q(ba) \gets Q(ba) + \frac{V - Q(ba)}{N(ba)}$
            \State $\vect{Q}_C(ba) \gets \vect{Q}_C(ba) + \frac{\vect{C} - \vect{Q}_C(ba)}{N(ba)}$
            \State $\bar{\vect{c}}(ba) \gets \bar{\vect{c}}(ba) + \frac{\vect{c} - \bar{\vect{c}}(ba)}{N(ba)}$
            \If{returnMinimalCost}
                \State $\vect{C} \gets \underset{a \in C(h)}{\min}\, \vect{Q}_C(ba)$
            \EndIf
            \State \textbf{return} $V, \vect{C}$
        \EndProcedure
    \end{algorithmic}
\end{algorithm}

Finally, we noted empirically that propagating simulated costs up the tree would results in overly conservative policies, as observation nodes with safe actions would propagate costs from unsafe actions up the tree, despite the fact that a closed-loop planner would choose the safe actions. To overcome this, in all of our solvers, we implement an option to propagate minimal costs across action nodes. With multiple constraints, this requires implementing a preference ordering to determine which cost value vector is returned.

\section{Experiments}

In this section, we consider three constrained variants of continuous POMDP planning problems in order to empirically demonstrate the efficacy of our methods. We compare the use of CPOMCPOW against POMCPOW with scalarized costs, the use of minimal cost propagation, and the use of different continuous CPOMDP solvers. 
We use Julia 1.6 and the POMDPs.jl framework in our experiments~\cite{egorov2017pomdps}. Our code is available at \url{https://github.com/sisl/CPOMDPExperiments}.

\subsubsection{CPOMDP Problems}

We enumerate the CPOMDP problems below, along with whether their state, action, and observation spaces are (D)iscrete or (C)ontinuous.
\begin{enumerate}
    \item \textbf{Constrained LightDark} (C, D, C): In this adaptation of LightDark \cite{sunberg2018online}, the agent can choose from moving in discrete steps of $\mathcal{A} = \{0, \pm 1, \pm 5, \pm 10\}$ in order to navigate to $s\in[-1,1]$, take action $0$, and receive $+100$ reward. An action of $0$ elsewhere accrues a $-100$ reward and the agent yields a per-step reward of $-1$. The agent starts in the dark region, $b_0 = \mathcal{N}(2,2)$, and can navigate towards the light region at $s=10$ to help localize itself. However, there is a cliff at $s=12$, above which the agent will receive a per-step cost of $1$. The agent must maintain a cost budget of $\hat{c}=0.1$, and so taking the $+10$ action immediately would violate the constraint.
    \item \textbf{Constrained Van Der Pol Tag} (C, C, C): In this problem a constant velocity agent must choose its orientation in order to intercept a partially observable target whose dynamics follow the Van der Pol oscillator~\cite{sunberg2018online}. In our adaptation, rather than penalizing taking good observations in the reward function, we formulate a cost constraint that dictates that the discounted number of good observations taken must be less than $2.5$.
    \item \textbf{Constrained Spillpoint} (C, C, C) This CPOMDP models safe geological carbon capture and sequestration around uncertain subsurface geometries and properties~\cite{spillpoint}. In the original POMDP, instances of CO$_2$ leaking through faults in the geometry are heavily penalized, both for the presence of a leak and for the total amount leaked. In our adaptation, we instead impose a hard constraint of no leaking.
\end{enumerate}

\begin{table*}[htpb]
\ra{1.2}
\caption{Continuous CPOMDP online algorithm results comparing mean discounted cumulative rewards and costs across 100 LightDark simulations, 50 Van Der Pol Tag simulations, and 10 Spillpoint simulations.}
\centering
\scalebox{0.82}{\begin{tabular}{@{}r RRRRRR @{}}
\toprule
&\multicolumn{2}{c}{\underline{\quad\quad\textbf{LightDark}\quad\quad}}
&\multicolumn{2}{c}{\underline{\quad\quad\textbf{Van Der Pol Tag}\quad\quad}}
&\multicolumn{2}{c}{\underline{\quad\quad\quad\textbf{Spillpoint}\quad\quad\quad}}\\
\textbf{Model} &  
{V_R}  & {V_C\ [\leq 0.1]} & 
{V_R}  & {V_C\ [\leq 2.5]} & 
{V_R}  & {V_C\ [\leq 0]}\\
\midrule
\texttt{CPOMCPOW}            & 
17.1 \scriptstyle\pm 0.7 & 0.090 \scriptstyle\pm 0.002  & 
\bf{24.5 \scriptstyle\pm 0.4} & 1.57 \scriptstyle\pm 0.001  &
3.93 \scriptstyle\pm 0.17 &0.001 \scriptstyle\pm 0.000  \\
\texttt{CPFT-DPW}            & 
\bf{51.9 \scriptstyle\pm 0.4} & 0.044 \scriptstyle\pm 0.002  & 
-0.6 \scriptstyle\pm 0.3 & 1.05 \scriptstyle\pm 0.01  &
\bf{4.19 \scriptstyle\pm 0.15} & 0.001 \scriptstyle\pm 0.000  \\
\texttt{CPOMCP-DPW}            & 
-6.5 \scriptstyle\pm 0.4 & 0.000 \scriptstyle\pm 0.000  & 
12.5 \scriptstyle\pm 0.5 & 1.71 \scriptstyle\pm 0.01  &
3.17 \scriptstyle\pm 0.18 & 0.001 \scriptstyle\pm 0.000  \\
\bottomrule
\end{tabular}}
\label{table:results}
\end{table*}

\subsubsection{Hard constraints vs. reward scalarization}
\begin{figure}[t]
\centering
\includegraphics[width=\columnwidth]{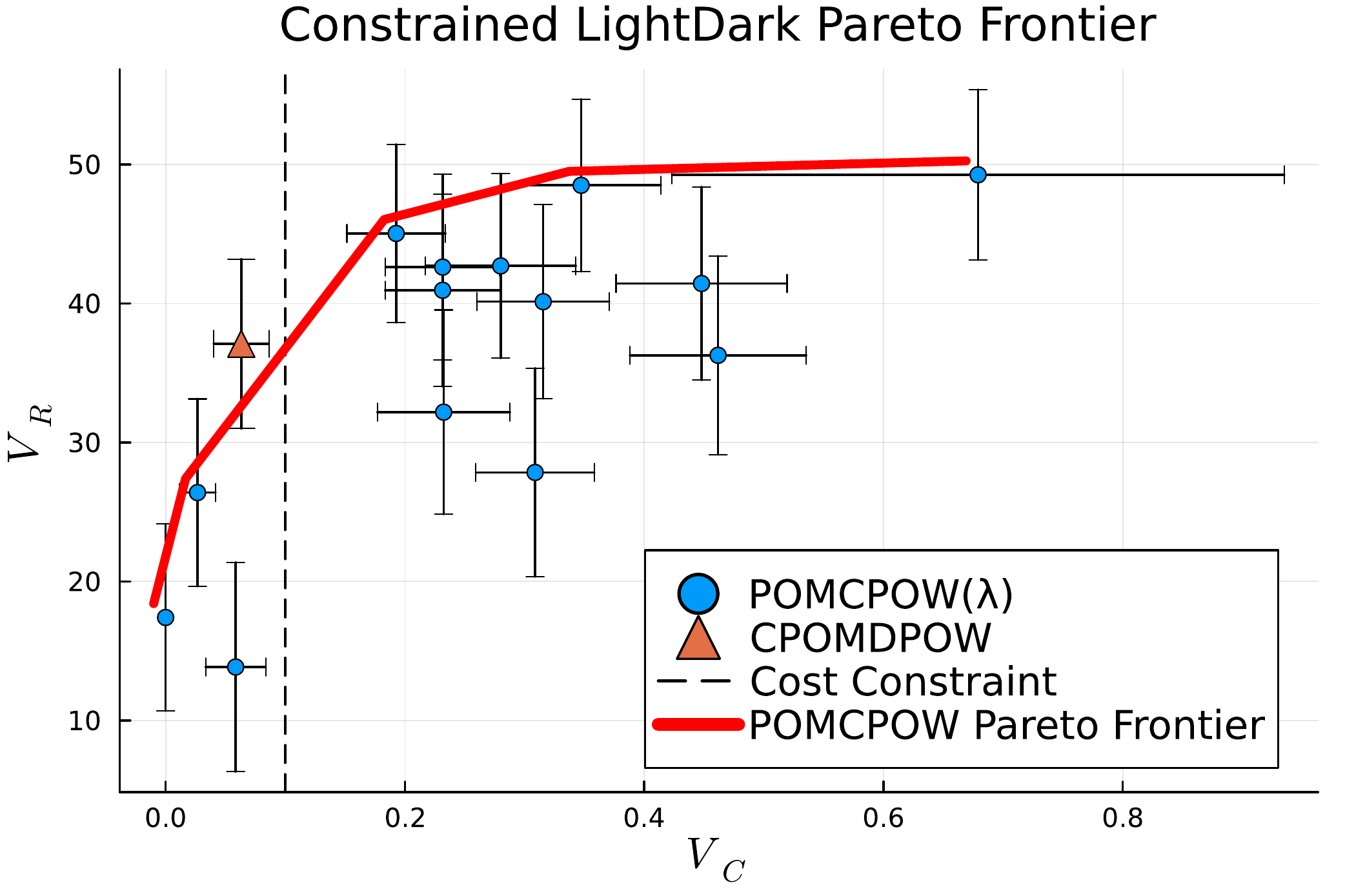} 
\caption{The $V_R$ vs. $V_C$ Pareto frontier of solutions to the scalarized LightDark POMDP solved with POMCPOW plotted on top of the Constrained LightDark CPOMDPOW solution. Error bars indicate standard deviation of the mean after 100 simulations.}
\label{fig1}
\end{figure}

First, we demonstrate the benefit of imposing hard constraints. To do so, we create an unconstrained LightDark problem by scalarizing the reward function and having an unconstrained solver optimize $\bar{R}(s,a) = R(s,a) - \lambda C(s,a)$. We then vary choices of $\lambda$ and compare the true reward and cost outcomes when using POMCPOW to solve the scalarized problem against using CPOMCPOW with the true CPOMDP. 

In \Cref{fig1}, we depict the Pareto frontier of the soft-constrained problem when simulating 100 episodes for each choice of $\lambda$. We see that the solution to the constrained problem using CPOMCPOW lies above the convex hull while consistently satisfying the cost constraint. We can therefore see that constrained solvers can yield a higher reward value at a fixed cost while eliminating the need to search over scalarization parameters.

\subsubsection{Cost propagation}
\begin{table}[htpb]
\ra{1.2}
\caption{Statistics corresponding with actions $1$, $5$, and $10$ when running CPOMCPOW with normal cost propagation, minimal cost propagation, and on the unconstrained problem. $\Delta Q_\lambda$ denotes the gap to the best Lagrangian action-value, and the action taken most often is highlighted in \textcolor{red}{red}.}
\centering
\scalebox{0.82}{\begin{tabular}{@{}l RRR @{}}
\toprule
\textbf{Model} &  
{N(b_0a)/N(b_0)} & {Q_c(b_0a)} & {\Delta Q_\lambda(b_0a)}   \\
\midrule
Normal           & 
[0.10, 0.19, 0.08] & [0.18, 0.32, 1.15]  & [-5.3, -5.5, -18.9] \\
Min            & 
[0.14, 0.33, 0.18] & [0.00, 0.00, 0.29]  & [-4.9, \textcolor{red}{-2.1}, -4.3] \\
\midrule
Uncstr.            & 
[0.09,0.30,0.35] & $---$  & [-8.9, -6.6, \textcolor{red}{-4.5}] \\
\bottomrule
\end{tabular}}
\label{table:costprop}
\end{table}

Additionally, we simulate 50 CPOMCPOW searches from the initial LightDark belief with and without minimal cost propagation. In Table 2, we compare average statistics for taking actions $1$, $5$, and $10$ after each search has been completed. While in the unconstrained problem, the agent chooses the $10$ action to localize itself quickly, we note that the constrained agent should choose the $5$ action to carefully move towards the light region without overshooting and violating the cost constraint. We see that with normal cost propagation, the costs at the top level of the search tree are overly pessimistic, noting that actions of $1$ or $5$ should have $0$ cost-value as they are recoverable. We see that with normal cost propagation, this pessimistic cost-value distributes the search towards overly conservative actions, while with minimal cost propagation, the search focuses around and picks the $5$ action.

\subsubsection{CPOMDP algorithm comparison}
\Cref{table:results} compares mean rewards and costs received using CPOMCPOW, CPFT-DPW, and CPOMCP-DPW  on the three target problems. 
Experimentation details are available in~\Cref{sec:hyperparameters}. 
Crucially, our methods can generate desirable behavior while satisfying hard constraints without scalarization. \textit{This is especially evident in the Spillpoint problem, where setting hard constraints minimizes CO$_2$ leakage while improving the reward generated by unconstrained POMCPOW reported by~\citet{spillpoint}. }

\section{Conclusion}

Planning under uncertainty is often multi-objective. Though multiple objectives can be scalarized into a single reward function with soft constraints, CPOMDPs provide a mathematical framework for POMDP planning with hard constraints. Previous work performs online CPOMDP planning for large state spaces, but small, discrete action and observation spaces by combining MCTS with dual ascent~\cite{lee2018monte}. We proposed algorithms that extend this to large or continuous action and observation spaces using progressive widening, demonstrating our solvers empirically on toy and real CPOMDP problems.

\textbf{Limitations} A significant drawback of CC-POMCP is that constraint violations are only satisfied in the limit, limiting its ability to be used as an anytime planner. This is worsened when actions and observations are continuous, as the progressive widening can miss subtrees of high cost. 
Finally, we note the limitation of using a single $\vect{\lambda}$ to guide the whole search as different belief nodes necessitate different safety considerations. 
\section*{Acknowledgments}

This material is based upon work supported by the National Science Foundation Graduate Research Fellowship Program under Grant No. DGE-1656518. Any opinions, findings, and conclusions or recommendations expressed in this material are those of the author(s) and do not necessarily reflect the views of the National Science Foundation. 
This work is also supported by the COMET K2---Competence Centers for Excellent Technologies Programme of the Federal Ministry for Transport, Innovation and Technology (bmvit), the Federal Ministry for Digital, Business and Enterprise (bmdw), the Austrian Research Promotion Agency (FFG), the Province of Styria, and the Styrian Business Promotion Agency (SFG). The authors would also like to acknowledge the funding support from OMV. We acknowledge Zachary Sunberg, John Mern, and Kyle Wray for insightful discussions.

\bibliography{sislstrings,references}
\newpage
\appendix

\renewcommand{\theequation}{A.\arabic{equation}}
\renewcommand{\thetable}{A.\arabic{table}}
\renewcommand{\thefigure}{A.\arabic{figure}}
\setcounter{equation}{0}
\setcounter{table}{0}
\setcounter{figure}{0}

\section{CPOMCP-DPW}
\begin{algorithm}
    \caption{CPOMCP-DPW} \label{alg:pomcpdpw}
    \begin{algorithmic}[1]
        \Procedure {Simulate}{$s$, $h$, $d$}        
            \If{$d = 0$}
                \State \textbf{return} $0$
            \EndIf
            \State $new \gets 0$
            \State $a \gets \Call{ActionProgWiden}{h}$
            \If{$|C(ha)| \leq k_o N(ha)^{\alpha_o}$}
                \State $s',o,r,\vect{c} \gets G(s,a)$
                \State $C(ha) \gets C(ha) \cup \{o\}$
                \State $M(hao) \gets M(hao) + 1$
                \State $\text{append } s' \text{ to } B(hao)$ \label{lin:insertion}
                \If{$M(hao) = 1$}
                    \State $new \gets 1$
                \EndIf
            \Else \label{lin:notnew}
                \State $o \gets \text{select } o \in C(ha) \text{ w.p. } \frac{M(hao)}{\sum_{o} M(hao)}$ \label{lin:selecto}
                \State $s' \gets \text{select } s' \in B(hao) \text{ w.p. } \frac{1}{|B(hao)|}$ \label{lin:samples}
                \State $r \gets R(s,a,s')$
                \State $\vect{c} \gets C(s,a,s')$
                
            \EndIf
            \If{$new$}
                \State $V', \vect{C}' \gets \Call{Rollout}{s', hao, d-1}$
            \Else
                \State $V', \vect{C}' \gets \Call{Simulate}{s', hao, d-1}$
            \EndIf
            \State $V \gets r + \gamma V'$
            \State $\vect{C} \gets \vect{c} + \gamma \vect{C}'$
            \State $N(h) \gets N(h)+1$
            \State $N(ha) \gets N(ha)+1$
            \State $Q(ha) \gets Q(ha) + \frac{V - Q(ha)}{N(ha)}$
            \State $\vect{Q}_C(ha) \gets \vect{Q}_C(ha) + \frac{\vect{C} - \vect{Q}_C(ha)}{N(ha)}$
            \State $\bar{\vect{c}}(ha) \gets \bar{\vect{c}}(ha) + \frac{\vect{c} - \bar{\vect{c}}(ha)}{N(ha)}$
            \If{returnMinimalCost}
                \State $\vect{C} \gets \underset{a \in C(h)}{\min}\, \vect{Q}_C(ha)$
            \EndIf
            \State \textbf{return} $V, \vect{C}$
            
        \EndProcedure
    \end{algorithmic}
\end{algorithm}
\Cref{alg:pomcpdpw} outlines the straightforward application of double progressive widening to CC-POMCP~\cite{lee2018monte}. As noted by~\citet{sunberg2018online}, this leads to rapid particle collapse in large or continuous observation spaces and performs similarly to a $Q_\text{MDP}$ policy, which assumes full observability in the next step. \label{sec:cpomcpdpw}

\section{Experimentation Details} \label{sec:hyperparameters}
Experimentation settings, model parameters, belief updaters, solver hyperparameters, and default rollout policies are visible on our codebase at \url{https://github.com/sisl/CPOMDPExperiments}. 
\end{document}